\documentstyle[11pt]{tmi}

\begin{document}

\title{A Lexicalist Approach to the Translation of Colloquial Text}

\author{ Fred Popowich\ \ Davide Turcato\ \ Olivier Laurens\ \ Paul
McFetridge}
\institute{Natural Language Lab\\
School of Computing Science\\
Simon Fraser University \\ 
Burnaby, BC \\
Canada V5A 1S6 \\
{\tt \{popowich,turk,laurens,mcfet\}@cs.sfu.ca}\\ 
\vspace{11pt}
{\normalsize J. Devlan Nicholson\ \ Patrick McGivern\ \ 
Maricela Corzo-Pena\ \ Lisa Pidruchney} \\
\vspace{11pt}
TCC Communications\\
Sidney, BC\\
Canada V8L 3Y3\\
{\tt \{devlan,patrickm,mcorzope,lpidruch\}@tcc.bc.ca}\\
\vspace{11pt}
{\normalsize Scott MacDonald} \\
\vspace{11pt}
Centre for Cognitive Science\\
2 Buccleuch Place\\
University of Edinburgh\\
Scotland EH8 9LW\\
{\tt scottm@cogsci.ed.ac}.uk\\
\vspace{11pt}
To appear in the proceedings of TMI-97, the 7th International Conference on 
Theoretical and Methodological Issues in Machine Translation. To be presented
at TMI-97, Santa Fe, New Mexico, July 23-25, 1997}

\markboth{To appear in the Proceedings of TMI-97}{To appear in the Proceedings of TMI-97}
\pagestyle{myheadings}
\maketitle

\begin{abstract}
  
  Colloquial English (CE) as found in television programs or typical
  conversations is different than text found in technical manuals,
  newspapers and books.  Phrases tend to be shorter and less
  sophisticated.  In this paper, we look at some of the theoretical
  and implementational issues involved in translating CE.  We present
  a fully automatic large-scale multilingual natural language
  processing system for translation of CE input text, as found in the
  commercially transmitted closed-caption television signal, into
  simple target sentences.  Our approach is based on the Whitelock's
  Shake and Bake machine translation paradigm, which relies heavily on
  lexical resources.  The system currently translates from English to
  Spanish with the translation modules for Brazilian Portuguese under
  development.
\end{abstract}

\section{Introduction.}
Colloquial language as found in typical conversations is different
than text found in technical manuals, newspapers and books. Phrases
tend to be shorter, less sophisticated, and may often contain
violations of linguistic rules and conventions found in written
language. 
A great deal of research in machine translation (MT) has focussed on the
translation of non-colloquial language.

We have tackled an MT task involving the translation
of Colloquial English (CE): the translation of the closed-caption text that
is transmitted with the vast majority of North American television
programs (and all major release videos).
The translation of closed-captions need not be as
accurate to be highly understandable if the translation is presented
simultaneously with the original television broadcast;
visual and contextual information from the broadcast can be used to supply
additional information, just as is done with the usual
presentation of the original closed caption text.

After examining the issues specific to the translation of CE as found
in closed captions, we shall describe a lexicalist approach which is
well suited for the translation task.  We will then introduce a fully
automatic large-scale multilingual natural language processing system
for translation of CE input text as found in many closed-caption
broadcasts, and examine issues relating to extension of the system for
handling additional languages.  The linguistic information is
explicitly represented in grammars and lexicons making up language
specific modules, so that modules for additional target languages can
be added in the future.  Spanish target language modules have already
been developed, with Brazilian Portuguese modules currently under
development.  Some results about the current system performance will
then be presented.

\section{\label{tcc}Translation of Closed Captions.}
The translation of closed captions is very peculiar and differs in
many ways from any other linguistic domain currently tackled by
machine translation systems. In this section, this peculiarity
will be described, with reference to the characteristics
of the language found in closed captions, and the conditions under which
the translated closed captions would be used.

CE sentences tend to be very short. Analysis of 11 million words of
text captured from typical North American prime-time television
broadcasts has shown the average sentence length to be 5.4 words, with
approximately 75\% of the sentences in the corpus having a length of 7
words or less and 90\% of the sentences consisting of 10 words
or less.  In many cases they are not actually complete sentences, but
phrasal fragments of some kind. Their syntax is considerably simpler
than that found in written text. Phrases tend to have a very simple
structure, subordination is rarely used and parataxis is often used
instead.  Elliptical expressions are frequently present.

CE contains a good deal of idiomatic expressions and slang, and is
frequently ungrammatical. Moreover, although closed captions are
written text, they are meant to render spoken language and therefore
they contain those phenomena which characterize spoken language:
hesitations, interruptions, repetitions, non linguistic utterances,
etc.  A taste of the kind of text available in closed captions is
provided by the fragment in Table \ref{script}, taken from the script
of the film {\em Copycat}.

\begin{table}
\begin{center}

\begin{tabular}{l}														\hline
{\tt morning.}					\\
{\tt morning, mj.}				\\
{\tt hi.}					\\
{\tt wow.}					\\
{\tt uh-huh.}					\\
{\tt pretty wild.}				\\
{\tt how you doing?}				\\
{\tt got an estimate for me?}			\\
{\tt about eight hours.}			\\
{\tt mm-hmm.}					\\
{\tt i am seeing ligature marks.}		\\
{\tt petechia.}					\\
{\tt strangled, huh?}				\\
{\tt same as the others.}			\\
{\tt yeah.}					\\
{\tt bag all the stuff on the bedside table?}	\\
{\tt thanks, bill.}				\\
{\tt that's the only one i need.}		\\
{\tt mike.}					\\
{\tt yes, madam.}				\\
{\tt were you the first one here?}		\\
{\tt yes.}					\\ \hline
\end{tabular}
    
\end{center}
\caption{\label{script}Script fragment}
\end{table}

Ambiguity is a further issue related to closed captions. On the one
hand, the shortness of phrases leaves little room for syntactic
ambiguity.  On the other hand, semantic ambiguity is often a
problem. The comprehension of a dialogue relies heavily on contextual
information not available in closed captions. Therefore, the meaning
of a sentence is sometimes difficult to recover. Reading the script of
a previously unseen movie is an instructive (and frustrating)
experience since it can be hard to understand what's happening and
what the characters are talking about, as is apparent from reading the
script fragment in Table \ref{script}.

Unlike translating technical manuals or weather reports, translating
TV programs means dealing with an unrestricted semantic domain. Since
we cannot currently provide the system with any of the current means
to deal with a semantic domain (world knowledge databases, example
based training, etc.), it is apparent that the system cannot currently
obtain a truly semantically accurate translation. Therefore, in the
traditional MT dilemma between {\em robustness} and {\em accuracy}, we
definitely favor the former over the latter. However, the disadvantage
of dealing with an input stripped away from its visual context is
countered by its symmetrical advantage that the user need not rely
only on the translated text in order to understand its meaning: users
can employ the visual context to make sense of the
translations. Therefore, translation acceptability to the end user
can be improved by the additional content and context provided by the
images and soundtrack.

In the light of the remarks above, the traditional requirement of {\em
meaning equivalence} between source and target sentence can be
weakened in the present context to a requirement of {\em meaning
subsumption}. As we have shown, the goal to achieve fully meaning
equivalent translations (i.e. target sentences that, ideally, can be
re-translated to their source sentences) is hardly achievable in the
present context, but fortunately is also less necessary than in other
MT domains. More modestly, our goal is to obtain an output consistent
with the input, in the sense that its content, although less specific
than the intended input meaning, must be coherent with it and hence
with its context. If such a goal is achieved, the user is likely to be
able to recover from the context the lesser information not found in the
translation.

Given this context, the traditional measure of accuracy of translation
can be replaced by a measure of acceptability of translation.
Sentences and phrases will be acceptable not only in the case where
they are accurate, but also in the case where they are understandable
by the viewer in the context of the simultaneous television broadcast.
Acceptability is thus defined in terms of the end user
\cite{Church:Crummy}.  Indeed, translations that contain one or two
minor errors (involving say lexical choice or modifier placement) will
often be acceptable to the end user in the final viewing context.

\section{Use of Lexicalist MT.}

After a short introduction to lexicalist MT, we will consider
the decisions which have been taken in
designing and implementing the system.
This will be followed by a detailed discussion of the specific system 
components, namely the lexicons and grammars.

\subsection{Overview of Lexicalist MT.}

Contemporary MT systems are usually classified as either interlingua based
or transfer based \cite{Hutchins:IMT}.  The lexicalist approach, or Shake and
Bake (S\&B) approach, \cite{Whitelock:SB,Beaven:SB} is similar to the
transfer-based approach in having distinct analysis, transfer
and generation stages. It differs though in the type of structures
that these three different stages use.

Many applications in machine translation, and in natural language
processing in general, make use of feature structures (also known as
attribute value matrices) for representing linguistic information (in
our case morphological, syntactic and semantic information) about
different words and phrases \cite{Carpenter:LTFS}.  A key aspect of
feature structures is that information can be shared between features,
and information can be initially unspecified (to be specified or
instantiated dynamically through unification).

The goal of the analysis phase in the S\&B approach 
is to produce a set of enriched lexical
entries associated with the different words in the source language
sentence. They are enriched in the sense that they contain additional
syntactic and semantic information above what was present in the
original lexical entries retrieved from the lexicon. During the
analysis phase, lexical entries get combined according to the grammar
rules and through unification values that were originally unspecified
in the feature structures from the lexicon become instantiated in the
feature structures in the parse tree. After analysis, these enriched
feature structures for the lexical entries are used as input to the
transfer module which together with lexical transfer rules, contained
in a bilingual (or multilingual) lexicon, result in the creation of a
corresponding set of feature structures for the target language. The
generator then uses this set of feature structures as input, together
with the target language lexicon and grammar to produce an output
sentence.

Typically, a transfer based system requires structural transfer rules
as well as lexical transfer rules. With the lexicalist approach, only
lexical transfer rules are needed. 
There is sometimes redundancy
between lexical transfer rules and structural transfer rules, so by
placing all the information in the lexical transfer rules, this
redundancy can be eliminated. 
Although we have no structural 
transfer rules, we still consider our approach to be transfer based 
since the lexical transfer rules are specific to a given source and
target language.
The lexicalist approach is also
attractive for translating idiomatic expressions, where the
translation of complex phrasal structures can be simply specified in
the transfer lexicon.
All the information relevant for translation will be contained in the
source language lexicon and grammar, the transfer lexicon, and the target
language lexicon and grammar.

\subsection{Why Lexicalist MT?}

A number of reasons suggested the use of a S\&B lexicalist approach
with respect to the specific domain addressed. As already pointed out,
colloquial text contains phrases of different sorts, rather than just
complete sentences, as is the case with most written texts (books,
newspapers, etc.). In the short sample provided in Table \ref{script}, 
noun phrases,
adjectival phrases, prepositional phrases, etc. can be found along
with sentences. One of the key features of the S\&B approach (and
other unification-based approaches to MT) is that
grammars can be developed in a declarative fashion, regardless of the
specific procedure for which they are going to be used. The purpose of
a grammar of this kind is to specify all the well-formed phrases of a
language by means of a set of constraints. A grammar of this kind
works equally well for full sentences as for any other kind of phrase.

We have also pointed out that colloquial expressions tend to be poorly
structured. Unlike written text, the nature of colloquial
communication is such that a speaker often starts an utterance before
having in mind a complete, structured sentence. As a consequence,
colloquial expressions often take the form of sequences of
unstructured, short phrases, rather than long sentences with complex
subordination. An MT system relying on structural transfer (thus
mapping ordered sequences onto ordered sequences) would have to assign
structures anyway to its source expressions, if these are to be
translated, mapping them onto corresponding target structures. This
would result in unnatural, artificially complex transfer mappings. In
the S\&B approach, although a structure is assigned in parsing,
structural information is stripped away from source
expressions. Transfer only maps lexical items. The only kind of
`structural' information transferred is a system of semantic
dependencies expressed by means of indices. However, such dependencies
can be specified in a partial way. They only need to be sufficiently
specified to avoid an incorrect swapping of elements of the same
category in generation. There is no need to assign and transfer this
sort of information, when it is irrelevant to translation. This
feature makes the translation process much easier when the input
contains elements whose position is largely accidental. For instance,
in the case of a parenthetical expression, which can appear
in virtually any position in the input, the transfer procedure need not take
into account all the possible positions where the expression can
appear. The expression is simply assigned no dependency, leaving it
free to appear in any position in the target sentence, provided that
the target grammar constraints are satisfied.

Further support for a lexicalist approach comes from the observation
that Colloquial English (and, more generally, the colloquial version
of any language) is richer in idiomatic expressions than written
English. Idiomatic expressions resist any compositional analysis and
can only be lexically translated as compound expressions. A lexicalist
approach, which is explicitly centered around a bilingual lexicon as a
means to perform lexical transfer, is therefore particularly suitable
for the expression of this sort of bilingual information.

\subsection{English Lexicon.}
Our first English lexicon was partially derived from the 70,000 entries of the machine
readable Oxford Advanced Learners Dictionary (OALD) which is rich in
morphological and syntactic information. Since English words do not
exhibit a great degree of inflectional variation, there is no need for
distinct morphological rules to be stored in the system. Instead, the
inflected form of the words can be explicitly stored in the lexicon
together with its associated morphological information. For example,
our English lexicon contains distinct entries for {\em dog} and {\em
  dogs}, one entry having information that it is a singular noun, and
the other that it is a plural noun. At the cost of the related
increase in storage requirements, we are able to avoid any English
morphological analysis and thus decrease our analysis time.

Our English lexicon also contains verb pattern information.
These patterns give detailed
information about the subcategorization requirements of the different
verbs (descriptions of the kinds of phrases and constituents that the
verbs can combine with). Rich subcategorization information is
precisely what is required by a lexical grammar formalism. The
relationship between the codes used in the OALD and the feature
structures used in our grammar is provided by a separate module. Thus,
if one wanted to simplify the system to ignore information supplied in
the OALD, or use some other machine readable dictionary other than the
OALD, only this mapping module would need to be modified.

\subsection{English Grammar.}
The initial use of the OALD has allowed us to implement an HPSG \cite{Pollard:HPSG}
style English grammar with syntactic information coded into lexical
entries, and constituency and order information contained in a small number of
grammar rules which gives us very broad coverage. 

The nature of colloquial text is such that the input to the machine
translation system may cover a very wide semantic domain, rather than
the more restrictive domains often found in technical text.  This
necessitates a shallow analysis of the input text, and a translation
process that is predominately based on morphological and syntactic
information with very little semantic information. In addition, there
is the need for a Colloquial English grammar as opposed to a more
formal English grammar.

A primitive notion of semantic roles is introduced into the grammar
and lexicon to allow semantic dependencies between constituents to be
represented, as is traditionally done in the S\&B approach. Each word
in a phrase or sentence is assigned an index, and this index can
appear as one of the semantic arguments of another word. For example,
in the sentence {\em John loves Mary}, the index of {\em John} would
be the first semantic argument of {\em loves} while the index of {\em
Mary} would be its second semantic argument. The ordering of the
arguments is determined according to their obliqueness, as is done in
HPSG -- the subject is the least oblique argument, followed by the
direct object, followed by the indirect object, etc.. It is also
possible for two words or phrases to share the same index. In this
case, they are said to be coindexed. Part of the role of the analysis
phase is to use the semantic indices to represent semantic
dependencies, which play a key role in the generation and transfer
phases.

The English grammar is used for shallow parsing. It is meant to be as
unrestrictive as possible, in order to accommodate for colloquial
expressions and constructions which are not strictly grammatical. It
is just restrictive enough though to augment the input lexical
entries with sufficient information for the subsequent transfer phase.
Any additional information which can be recovered in some other way
during transfer is superfluous and thus avoided. For instance, the
English grammar does not enforce agreement between subject and verb,
because the Spanish grammar still has the means (by reference to argument
indices) to recognize
the subject-verb relation and enforce proper agreement on the Spanish
output. Although the grammar is purely declarative and thus usable in
principle in both directions, its unrestrictive nature makes it specifically
suitable as a source language grammar.

\subsection{Bilingual Lexicon.}
The bilingual lexicon is responsible for the transfer aspect in an
S\&B system.  Although S\&B permits bidirectional translation, we can
take advantage of constraints on transfer rules by building a
unidirectional dictionary and obtain better performance. Each entry in
the bilingual lexicon establishes a relationship between a set of
source language lexemes and a set of target language lexemes.

\subsubsection{Bilingual entries.}
In our English-Spanish bilingual lexicon many of these entries are one
to one --- a single feature structure for an English word is mapped to a
single Spanish feature structure. Information from the source language
feature structure may be related to information in the target language
feature structure. For example, this information may include the value
of agreement features such as person or number, tense and semantic
indices.

Perhaps more interesting are the bilingual entries which involve
multiple lexemes on either side, e.g. {\em `loud and clear}'
$\leftrightarrow$ {\em `perfectamente'} (many to one), {\em `trip'}
$\leftrightarrow$ {\em `hacer tropezar'} (one to many), {\em `all over
  the place'} $\leftrightarrow$ {\em `por todas partes'} (many to
many). Such entries allow the mapping of English phrases to Spanish
phrases in the bilingual lexicon. This is important for colloquial
speech since idioms are commonly used. Idioms rarely translate
compositionally.  e.g. {\em `kick the bucket'} $\leftrightarrow$ {\em
  `estirar la pata'}, {\em `to be bending over backwards'}
$\leftrightarrow$ {\em `hacer lo imposible'}.

Expressing collocations in the entries can also help reduce ambiguity
(especially in the absence of deep semantic analysis), e.g. the
translation of {\em get} ({\em `get lost'}, {\em `get on with'}, {\em
  `get away from'}, {\em `get away with'}, {\em `get drunk'} \ldots)
or support verbs like {\em make} ({\em `make a phone call'}, {\em
  `make a decision'}, {\em `make a move'}, {\em `make trouble'}
\ldots).

It is important to note that the English phrase contained in the
bilingual entry need not be a single constituent in the English
sentence, e.g {\em `put up with'} $\leftrightarrow$ {\em `aguantar'}
where {\em with} is part of another constituent, a PP, the rest of
which is transferred by other entries. So the presence of additional
words within the phrase contained in the English sentence will not
interfere with the application of the rule. Thus the same rule can be
used in a wide variety of contexts.

Now, at a more detailed level,
a bilingual entry (transfer rule) is made up of the following
elements: a key-word, a source side, a target side, associated
transfer macros ({\em t-macros}). Each side (source or target) contains one
or more lexical signs (or even zero for the target side). The key-word
corresponds to the base form of one of the lexical signs on the source
side.

A transfer macro is basically an additional transfer rule called by
the transfer rule to which it is associated. Instead of a key-word a
transfer macro contains a description which the calling transfer rules
must satisfy in order for the transfer macro to be triggered. 

\subsubsection{Transfer macros.}
The use of transfer macros has proven to be useful in porting the
system to new language pairs. The purpose of a transfer macro is to
state some general relationship between structures in the source and
target language. In this sense, its purpose is similar to that of
transfer rules in traditional transfer approaches, with the crucial
advantage that, in the case of transfer macros, no structural mapping
is performed. The ability of a transfer macro to express generalizations
over pairs of languages in a compact form, and their systematic use in
this view, makes it possible to port a bilingual lexicon to a new
language pair without affecting the actual lexical entries (except
their target side, of course). Only transfer macros need to be
modified in order to accommodate the generalizations holding for the
new language pair.

To give an example, a generalization for the English-Spanish pair is
that a nominal indirect object (e.g. {\em `I tell the man that
\ldots'}), is mapped to a prepositional phrase plus a redundant
personal pronoun ({\em `Yo le digo al hombre que \ldots'}).  Such
complex transfer is performed in the English-Spanish bilingual lexicon
by means of a specific transfer macro associated with verbs taking a
dative complement. In porting the bilingual lexicon to
English-Brazilian Portuguese ({\em `Eu digo ao homem que \ldots'}) it
was sufficient to change the specific transfer macro, removing the
redundant pronoun requirement, without modifying anything from the
actual transfer rules apart from the right hand side words.

\subsection{Spanish Lexicon.}
The Spanish lexicon is converted from a Spanish lexicon used in the
METAL MT system \cite{Bennett:METAL}, containing about 25,000
entries. In addition to syntactic information, the lexicon contains
semantic features for nouns. Verb entries contain detailed
subcategorization information, along with semantic selectional
restrictions on nominal complements.

\subsection{Spanish Grammar.}
The Spanish grammar has been developed in the
spirit of traditional Phrase Structure Grammars, since it doesn't make
use of any Subcategorization Principle or Head Feature Principle. No
X-bar system is in use and the relation between lexical signs and
their phrasal projections is obtained by explicit stipulation in each
rule. Subcategorization is accomplished in a GPSG style \cite{Gazdar:GPSG}, 
with as many
different rules as subcategorization frames and an atomic-valued
{\tt subcat} feature which allows the match between specific verbs and
appropriate rules.

\subsubsection{Robustness vs. restrictiveness.}
The source and target grammars have been developed using the same
declarative formalism and can be used in principle for both parsing
and generation. However, although there is no commitment to a specific
use from a formal point of view, the bias towards a specific use has
played a role in their development, suggesting different guidelines in
view of different goals. While the source grammar has been developed
in view of robustness, the target grammar is meant to be highly
constrained and to generate only strictly grammatical output. The
generative power of the source grammar is actually a superset of
strictly grammatical English, whereas the generative power of the
target grammar is a subset of Spanish. A grammar for generation can be
fully adequate without generating every possible sentence in the
target language, provided that, for every sentence ruled out, there is
at least one equivalent sentence that can be generated. Such an
assumption makes things easier, for instance, when dealing with
languages with a high degree of word order freedom.

\subsubsection{Efficiency.}
The development of the target grammar has also followed the guideline
to have many, very specific rules rather than few, highly general rule
schemata. Such guidelines rest upon the assumption that simpler
lexical signs and highly specified rules reduce the amount of
unification performed in the generation process and reduce the number of
active edges in a chart. Grammar rules are very specific because signs
in rules are always specified for their category. Therefore
subcategorization and head feature percolation are obtained by means
of explicit stipulation rather than unification between daughter signs
and between head daughter and mother signs.

\subsubsection{Portability to new languages.}
Although the grammar rules are very specific, some effort has been
made to make a distinction between language specific, idiosyncratic
constraints and linguistic generalizations valid across different
languages. The latter kind of information has been stored in the body
of the rules themselves, while the former has been stored in goals
associated to the rules. A consistent use of such distinction has made
easier the porting of many rules across different languages, namely
Spanish and Portuguese. Many of the linguistic generalizations stated
in the rules could be preserved across the two languages, while the
replacement or modifications of associated goals made possible to
accommodate for different language specific constraints.

\section{Performance.}

Although the system is still under development, we have already
obtained some interesting results concerning its performance.  Table
\ref{accuracy} shows the result of a test on a collection of two
hundred random sentences from TV program scripts. The grammars and
lexicons were sufficiently developed to allow 84.5\% of the sentences
to be translated using the S\&B approach.  The translations were
evaluated by a native Spanish speaker as either correct ({\bf yes}),
acceptable though not ideal ({\bf ok}), or unacceptable ({\bf no}).
Approximately two thirds of the sentences translated are
understandable.

\begin{table}
  \begin{center}

  \begin{tabular}{lc}						\hline
	Evaluation\hspace{.3in}	& Shake and Bake	\\	\hline
	yes			& 38.46			\\
	ok			& 27.81			\\ 
	no			& 33.73			\\ 
	yes/ok			& 66.27			\\	\hline
  \end{tabular}
    
  \end{center}
\caption{\label{accuracy}Accuracy results}
\end{table}

Table \ref{sample} shows the system output for the short movie script
fragment from Table \ref{script} in section \ref{tcc}, along with
evaluations, as described above. 
Notice that three of the sentence from Table \ref{script} could not
be translated by the current state of the S\&B system.
The translation sample exemplifies
some features of the system (along with some problems):

\begin{itemize}
\item The correct translation of `{\tt how you doing?}' shows that the
system is able to associate grammatical translations to ungrammatical
input. A shallow parse is performed, which accepts and assigns a
structure to ill-formed input. However, the information added to the
source bag by the parsing step is sufficient to enable a mapping
onto a correct target bag in the transfer step.

\item As an example of how the visual context can contribute to
increase the acceptability of an incorrect translation, let's consider
the following translation:

\begin{quote}
{\tt strangled, huh?} $\Rightarrow$ {\tt ¿estrangul\'{o}, eh?}
\end{quote}

From a purely semantic point of view, there is a remarkable difference
between the English source and the Spanish translation: a passive
participle, `{\tt strangled}', is turned into an active verb, `{\tt
estrangul\'{o}}'. However, if the visual context makes clear that the
anaphora referent of the sentence is a corpse lying on the floor, the
user is very likely to have a clue for the correct interpretation of
the sentence.

\item The sample shows how frequent parenthetical expressions and
vocative forms are (`{\tt morning, mj.}', `{\tt strangled, huh?}',
`{\tt yes, madam.}'). 
It is the matter of the target grammar to place
the translation of the parenthetical expression at the right place,
according to purely monolingual constraints.

\end{itemize}

\begin{table}
  \begin{center}

  \begin{tabular}{llc}														\hline
English						& Spanish						& Eval.		\\	\hline
{\tt morning.}					& {\tt buenos d\'{i}as.}				& yes		\\
{\tt morning, mj.}				& {\tt buenos d\'{i}as, [mj].}				& yes		\\
{\tt hi.}					& {\tt hola.}						& yes		\\
{\tt wow.}					& {\tt fant\'{a}stico.}					& yes		\\
{\tt uh-huh.}					& {\tt [uh-huh].}					& no		\\
{\tt pretty wild.}				& {\tt bastante alocado.}				& yes		\\
{\tt how you doing?}				& {\tt ¿qu\'{e} tal est\'{a}?}				& yes		\\
{\tt got an estimate for me?}			& {\tt ¿recibido un c\'{a}lculo para m\'{i}?}		& ok		\\
{\tt about eight hours.}			& {\tt de ocho horas.}					& no		\\
{\tt mm-hmm.}					& {\tt [mm-hmm].}					& ok		\\
{\tt i am seeing ligature marks.}		& {\tt veo las marcas de la ligadura.}			& yes		\\
{\tt petechia.}					& {\tt [petechia].}					& no		\\
{\tt strangled, huh?}				& {\tt ¿estrangul\'{o}, eh?}				& ok		\\
{\tt same as the others.}			& {\tt mismo como los otros.}				& ok		\\
{\tt yeah.}					& {\tt s\'{i}.}						& yes		\\
{\tt bag all the stuff on the bedside table?}	& 							& 		\\
{\tt thanks, bill.}				& {\tt gracias, [Bill].}				& yes		\\
{\tt that's the only one i need.}		& 							& 		\\
{\tt mike.}					& {\tt [Mike].}						& yes		\\
{\tt yes, madam.}				& {\tt s\'{i}, se\~{n}ora.}				& yes		\\
{\tt were you the first one here?}		&							&   		\\
{\tt yes.}					& {\tt s\'{i}.}						& yes		\\	\hline
  \end{tabular}
    
  \end{center}
\caption{\label{sample}Translation sample}
\end{table}

\section{Conclusion.}

The S\&B approach is very suitable for our specific domain.
It allows a simple treatment of idioms and other multiword constructions, 
it does not require a deep semantic analysis of expressions,  
it requires only lexical transfer
rules, and it is highly modular allowing rapid extension to new
target languages.
In addition to these advantages, the peculiarities of
CE allow us to avoid the computational complexity problems associated
with the S\&B approach \cite{Brew:COLING92};
a high percentage of our input is made
of very short sentences, so the computation associated with generation is
kept within acceptable limits \cite{Popowich:Generation}.

Church and Hovy have emphasized the importance of selecting
an appropriate application for MT, and the case could be made for NLP
in general \cite{Church:Crummy}. Inflated expectations can only result in 
disillusionment.
Automatic translation of CE seems appropriate for the current state
of the technology in appropriate environments. These environments
have the features of either augmenting the information source with other
input signals (as in the translation of closed captions) or providing the
user with time to acclimatize to the system and to decipher inappropriate
translations (as in the translation of typed conversation over the Internet).

The constraints on these applications -- the robustness requirements
and shallow analysis -- as well as the practical need for rapid
development of new language modules are adequately served by the
lexicalist approach.  Plans are in place for the rapid development of new
language modules for Italian, French and German.

\section*{Acknowledgements.}

The authors would like to thank Dan Fass, John Grayson and the anonymous referees
for their suggestions and observations.

\end{document}